# AI-Driven Prognostics for State of Health Prediction in Li-ion Batteries: A Comprehensive Analysis with Validation

Tianqi(Kirk) Ding
De*pt. of Electrical and Computer Engineering*
Baylor University
Waco, TX USA
Kirk_Ding1@baylor.edu

Dawei Xiang
De*pt. of Computer Science and Engineering*
University of Connecticut
Storrs, CT USA
ieb24002@uconn.edu

Tianyao Sun
Independent Researcher
New York, NY USA
sunstella313@gmail.com

YiJiashun Qi
De*pt. of Electrical and Computer Engineering*
University of Michigan
Ann Arbor, MI USA
elijahqi@umich.edu

Zunduo Zhao*
De*pt. of Computer Science*
New York University
New York, NY USA
Corresponding Author:
zz300@nyu.edu

*Abstract* — **This paper presents a comprehensive review of AI-driven prognostics for State of Health (SoH) prediction in lithium-ion batteries. We compare the effectiveness of various AI algorithms, including FFNN, LSTM, and BiLSTM, across multiple datasets (CALCE, NASA, UDDS) and scenarios (e.g., varying temperatures and driving conditions). Additionally, we analyze the factors influencing SoH fluctuations, such as temperature and charge-discharge rates, and validate our findings through simulations. The results demonstrate that BiLSTM achieves the highest accuracy, with an average RMSE reduction of 15% compared to LSTM, highlighting its robustness in real-world applications.**

*Keywords— lithium-ion battery, state of health (SoH) prediction, Electric Vehicle, Artificial Intelligence*

## I. Introduction

The global transportation sector contributes approximately 25% of the world's annual carbon emissions [1]. With the increasing concerns over climate change and air pollution, the transition from fossil-fuel-based vehicles to electric vehicles (EVs) has become a crucial step in achieving a sustainable and low-carbon future [2]. Governments worldwide have implemented policies and incentives to promote the adoption of EVs, leading to exponential growth in EV sales over the past decade. However, as the adoption of EVs continues to expand, the demand for high-performance, long-lasting, and reliable battery systems has also increased significantly.

Among various battery technologies, lithium-ion batteries (Li-ion batteries) have emerged as the dominant choice for EV applications due to their high energy density, lightweight properties, and declining production costs [3]. Despite these advantages, Li-ion batteries face a major challenge: capacity degradation over time, which directly impacts vehicle performance, driving range, and overall battery lifespan. Battery degradation is influenced by several factors, including temperature, charging cycles, and discharge rates [4]. Therefore, accurately predicting the State of Health (SoH) of Li-ion batteries is essential for battery safety, maintenance planning, and efficient energy management in EVs [5].

The Battery Management System (BMS) plays a critical role in monitoring and managing battery performance by assessing State of Charge (SoC) and State of Health (SoH) metrics [6]. SoH estimation helps determine the remaining useful life (RUL) of a battery and provides early warnings for battery replacements or safety hazards. Traditional methods for SoH estimation are primarily experiment-based or model-based. Experiment-based methods rely on extensive laboratory testing under controlled conditions, which are often costly and time-consuming. Model-based methods, on the other hand, employ electrochemical, statistical, or machine learning models to predict battery degradation patterns. However, these models often suffer from limited generalizability and lower accuracy when applied to real-world EV conditions.

With the rapid advancements in Artificial Intelligence (AI) and machine learning (ML) techniques, AI-driven SoH prediction models have gained increasing attention in recent years [5]. AI algorithms, particularly Recurrent Neural Networks (RNNs), Long Short-Term Memory (LSTM) networks, and Feed-Forward Neural Networks (FFNNs), have demonstrated superior performance in time-series analysis and pattern recognition, making them well-suited for battery SoH estimation [7]. These AI-based models can learn complex nonlinear relationships between input features (such as voltage, current, and temperature) and battery degradation trends, enabling more accurate and real-time predictions [8].

This paper provides a comprehensive review of AI applications in Li-ion battery SoH estimation. We summarize and compare existing AI-based prediction models, evaluate their accuracy and computational efficiency, and discuss the key challenges associated with their implementation in EV battery systems

## II. BACKGROUND

Battery health prediction plays a crucial role in the efficient management of lithium-ion (Li-ion) batteries in electric vehicles (EVs). The accurate estimation of a battery's State of Health (SoH) is essential for ensuring vehicle reliability, preventing unexpected failures, and optimizing battery life. Traditional SoH estimation methods rely on physical models and empirical experiments, but these approaches often suffer from high computational costs and limited adaptability to real-world scenarios. With the rapid advancement of Artificial Intelligence (AI) and machine learning (ML) techniques, data-driven models have emerged as a promising alternative for SoH prediction. This section introduces the fundamentals of Battery Management Systems (BMS), AI-based SoH prediction methods, and widely used battery datasets in the research community.

### A. Lithium-ion Battery Management System (BMS)

A BMS continuously monitors electrical, thermal, and chemical parameters to maintain optimal battery performance. Key measurements include:

- **Voltage Measurement**: Monitors the voltage of each individual cell and battery pack to prevent overvoltage or undervoltage conditions, which can lead to premature degradation or failure.
- **Current Measurement**: Tracks charging and discharging currents to manage power flow and prevent excessive current draw, which may cause overheating.
- **Temperature Measurement**: Utilizes thermistors or temperature sensors to detect overheating, preventing thermal runaway—a critical safety hazard.
- **State of Charge (SoC) Estimation**: Determines the remaining usable energy in the battery by integrating voltage, current, and temperature data.
- **State of Health (SoH) Estimation**: Evaluates battery degradation and aging over time to predict its remaining useful life (RUL). SoH is typically defined by the equation:

$$SOH = \frac{C_{recent}}{C_{initial}} \times 100 (\%) \quad (1)$$

$C_{Current}$ is the current capacity of the battery and $C_{initial}$ is the initial capacity of the battery [7]. The threshold value for SoH is 80%. According to IEEE Std. 1188.1996, when the SoH value of a lithium-ion battery is less than or equal to 80%, the Electric Vehicle needs to replace the battery [9].

Accurate SoH can be used to estimate SoC. SoC is the percentage of the battery's current remaining charge expressed as a ratio to its rated capacity. It is often used to prevent deep discharging or overcharging of a battery [10].

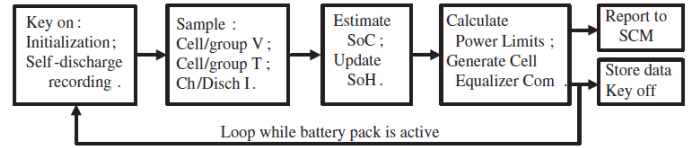

Fig 1. Parameter estimations and pack management in a BMS[10]

**2) Battery Protection and Fault Diagnosis**

Li-ion batteries are highly sensitive to overcharging, deep discharging, overheating, and short circuits, which can lead to serious safety hazards. A BMS integrates multiple safety mechanisms to prevent critical failures:

- **Overvoltage and Undervoltage Protection**:
  - Overcharging (>4.2V per cell) leads to lithium plating, which reduces battery life and increases the risk of short circuits.
  - Deep discharging (<2.5V per cell) accelerates irreversible capacity loss.
  - The BMS disconnects the charger or load when unsafe voltage levels are detected.
- **Overcurrent Protection**:
  - Excessive current flow during fast charging or sudden acceleration generates heat and stresses battery materials.
  - The BMS limits maximum charge/discharge currents and trips a safety cutoff switch if excessive currents are detected.
- **Thermal Management & Overtemperature Protection**:
  - High temperatures (>60°C) accelerate battery aging and increase the risk of thermal runaway.
  - The BMS activates cooling mechanisms (air cooling, liquid cooling, or phase-change materials) to regulate battery temperature.
- **Short-Circuit Protection**:
  - Internal or external short circuits can result in immediate catastrophic failure.
  - High-speed electronic fuses (MOSFETs or IGBTs) disconnect the circuit within milliseconds.
- **Cell Balancing**:
  - Over time, individual cells in a battery pack develop voltage imbalances, leading to reduced overall capacity.
  - BMS implements active balancing (energy transfer between cells) or passive

balancing (energy dissipation via resistors) to maintain uniform voltage levels.

**3) SoC and SoH Estimation Algorithms**

Since Li-ion batteries exhibit **highly nonlinear charge-discharge characteristics**, direct measurement of **SoC and SoH** is challenging [11]. Instead, the BMS employs **advanced estimation algorithms** to improve accuracy:

**SoC Estimation Methods**

- **Coulomb Counting Method**: Measures charge input/output but suffers from cumulative drift errors.

- **Open Circuit Voltage (OCV) Method**: Maps OCV to SoC but requires the battery to be in rest mode.

- **Kalman Filter (KF) & Extended Kalman Filter (EKF)**: Dynamically estimates SoC using a state-space **model**, reducing errors.

- **Artificial Neural Networks (ANNs)**: AI-based models trained on historical battery data provide more accurate SoC predictions.

**SoH Estimation Methods**

- **Capacity Fade Method**: Measures degradation over charge cycles.

- **Internal Resistance Measurement**: Tracks increasing internal resistance as an indicator of aging.

- **Machine Learning & Deep Learning Models**: Use large datasets to predict battery health trends with **higher precision**.

*B. Artificial Intelligence Algorithm*

Currently, EV lithium-ion battery SoH estimation methods fall into **two main categories**:

- **Experiment-based methods**: Conduct direct physical measurements under controlled laboratory conditions, offering high accuracy but requiring significant time and resources.
- **Model-based methods**: Use **physics-based, statistical, or AI-driven** approaches to predict SoH under real-world conditions (as shown in Fig. 2) [12].

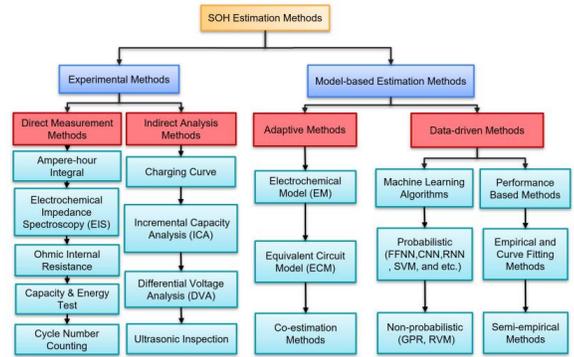

Fig 2. Methods of estimating SOH of battery [12]

Due to the **nonlinear characteristics** of battery degradation and the increasing demand for real-time estimation, AI-based **model-driven methods** have become the dominant approach for SoH prediction. The most commonly used AI algorithms include:

a. RNN (Recurrent Neural Network): is a neural network structure with recurrent connections, which is widely used in tasks such as natural language processing, speech recognition, and time-series data analysis. Compared to traditional neural networks, the main feature of RNN is that it can handle sequential data and can capture the temporal information in the sequence [13-16]. To put it simply: my favorite university is Baylor University, and I will definitely apply to xx. Here xx can be inferred to be Baylor University based on the context, but it is difficult for neural networks to do this, because neural network prerequisite inputs and outputs are also independent. So the improvement of RNN is to make the neural network with memory, so the output of the network can depend on the current input and memory [17].

From the RNN flowchart(shown in Fig. 3) at the bottom we can see that in an RNN network the hidden layer is used to store previous input information.

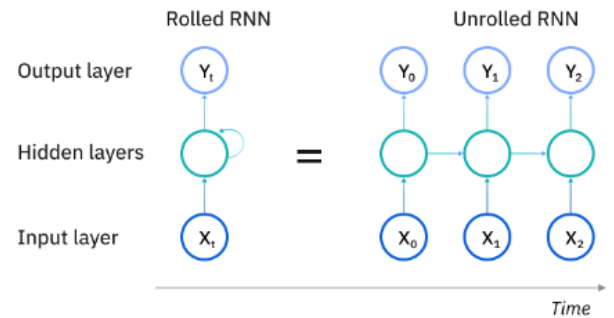

Fig 3. RNN Flowchart [18]

b. LSTM(Long short-term memory):

LSTM is a special kind of RNN, but compared with RNN, it can solve the problem of gradient disappearance (for RNN, for each time point, the information in memory will be covered, but LSTM is different, it is to multiply the value of the original memory by the value of the input and put it into the cell, so that the memory and input are summed up,(shown in Fig. 4) so unlike RNN, the information will be covered at each time point, as long as the information of the previous moment is formatted, the effect will disappear, but in LSTM, the effect will always be there. The memory and input are summed up, so unlike RNN which is overwritten at every point in time, as soon as the information of the previous moment is formatted, the effect disappears, but in LSTM the effect is always there) [19] [20].

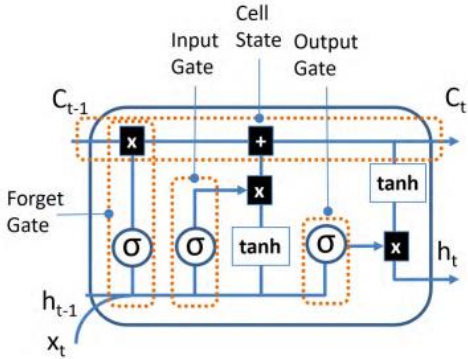

Fig 4. LSTM block diagram [18]

c. FFNN(Feed Forward Neural Networks): The signal flows unidirectionally, from the input layer to the hidden layers and then to the output layer, without feedback loops, and it is suitable for non-time series prediction problems.(Shown in Fig. 5) In time series prediction, feed forward neural network (FFNN) Although there is no time dependency, it can be used to realize the function of time series prediction by appropriate data preprocessing.

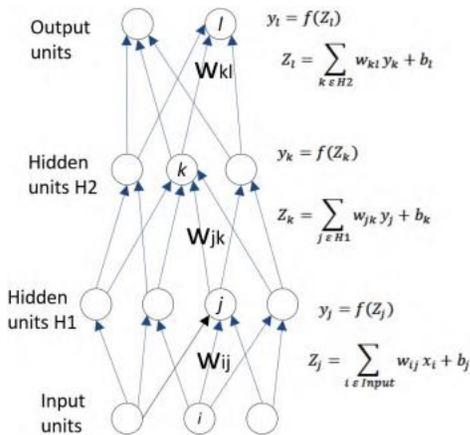

Fig 5. Feed Forward Neural Network [18]

C. Model Training Dataset

Any AI model requires data input. Currently the following datasets are commonly used by researchers for SoH prediction of EV. three datasets are CALCE(Center for Advanced Life Cycle Engineering, Collected by CALCE Battery Team in University of Maryland) [21], NASA(collected by NASA Ames Prognostics Center of Excellence) [19] and UDDS(Urban Dynamometer Driving Schedule, Collected by Dr. Phillip Kollmeyer at University of Wisconsin-Madison) [22].

III. EXPERIMENTS

Because each article method is very different however, it will be slightly similar, but due to the limited open source datasets available online. It is basically the 3 datasets provided above. Here we will categorize all experiments by using the same datasets.

A. CALCE

The researchers used data from the CALCE dataset covering 13 lithium-ion battery cells tested in series at temperatures from 20°C to 60°C and utilized the FFNN network to identify and classify them in order to calculate the time and history of the battery loads and thus make SoH predictions. The experimenters did two separate experiments. The error rates of the two experiments were 1) 0.6% for the first one and 2) 1.3% for the second one.(Shown in Fig. 6 a&b) [23]

| SOH output class [-] | 100-95 | 95-90 | 90-85 | 85-80 | 80-75 | |
|---|---|---|---|---|---|---|
| 100-95 | 23962 / 24.4% | 137 / 0.1% | 3 / 0.0% | 0 / 0.0% | 0 / 0.0% | 99.4% / 0.6% |
| 95-90 | 69 / 0.1% | 16235 / 16.5% | 92 / 0.1% | 0 / 0.0% | 0 / 0.0% | 99.0% / 1.0% |
| 90-85 | 29 / 0.0% | 61 / 0.1% | 17896 / 18.2% | 80 / 0.1% | 0 / 0.0% | 99.1% / 0.9% |
| 85-80 | 0 / 0.0% | 8 / 0.0% | 54 / 0.1% | 17923 / 18.2% | 47 / 0.0% | 99.4% / 0.6% |
| 80-75 | 0 / 0.0% | 0 / 0.0% | 0 / 0.0% | 42 / 0.0% | 21607 / 22.0% | 99.8% / 0.2% |
| | 99.6% / 0.4% | 98.7% / 1.3% | 99.2% / 0.8% | 99.3% / 0.7% | 99.8% / 0.2% | 99.4% / 0.6% |

SOH target class [-]

Fig 6a. Results of SOH accuracy [24]

Fig 6b. Results of SOH accuracy[17]

## B. NASA

The researchers used battery #5 (Li-Ion 18650) from the NASA battery pack to test data values at 25° ambient temperature. The experiment was divided into two parts a single channel (voltage only) SoH value prediction and multi-channel data (voltage, current, temperature) SoH value prediction. The experimenters used three neural networks FFNN, CNN (Convolutional Neural Network), LSTM for training and then the correctness of these three networks were compared in each section. Both single-channel and multi-channel results responded that LSTM has higher accuracy.(Shown in Fig. 7a&b) [25]

| Model | RMSE | MAE | MAPE(%) |
|---|---|---|---|
| FNN-1 (baseline) | 0.0736 | 0.0655 | 4.7100 |
| FNN-2 | 0.0633 | 0.0557 | 3.6500 |
| CNN-1 | 0.0701 | 0.0623 | 4.0020 |
| CNN-2 | 0.0766 | 0.0687 | 4.4187 |
| LSTM | 0.0288 | 0.0210 | 1.3770 |

Fig 7a. Signal Channel Estimation Errors

| Model | RMSE | MAE | MAPE(%) |
|---|---|---|---|
| MC-FNN-1 | 0.0379 | 0.0329 | 1.9800 |
| MC-FNN-2 | 0.0298 | 0.0242 | 1.7300 |
| MC-CNN-1 | 0.0584 | 0.0443 | 2.8961 |
| MC-CNN-2 | 0.0443 | 0.0364 | 2.3731 |
| MC-LSTM | 0.0246 | 0.0159 | 1.0320 |

Fig 7b. Multi-Channel Estimation Errors

The experimenters used charge and discharge data from four cells, #5, #6, #7, and #18, from the NASA battery pack. Since it is the same data set, these data were all measured at the 25° test ambient temperature. The experiment, like the one above, was also divided into two parts a single channel (voltage only) SoH value prediction as well as multi-channel data (voltage, current, temperature) SoH value prediction. The experimenters used six artificial intelligence algorithms: SVP (Support Vector Regression), AB (Adaptive Boosting), AB (Adaptive Boosting), MLP (Multi-Layer Perceptron), CNN, LSTM and BiLSTM (Bi-Directional Long Short-Term Memory Network). Long Short-Term Memory Network), and then each section compares the correctness of these six networks [24]. Both single-channel and multi-channel results reflect that BiLSTM has higher accuracy.(Shown in Fig. 8 a&b) [26]

| Methods | MSE | RMSE | MAE | MAPE |
|---|---|---|---|---|
| **Battery #5** | | | | |
| SVR | 0.8775 | 0.9368 | 0.8991 | 0.9253 |
| AB | 0.2889 | 0.5375 | 0.4272 | 0.6878 |
| MLP | 0.0011 | 0.0335 | 0.0268 | 0.0498 |
| CNN | 0.0024 | 0.0487 | 0.0319 | 0.0519 |
| LSTM | 0.0006 | 0.0256 | 0.0167 | 0.0372 |
| **BiLSTM** | **0.0001** | **0.0100** | **0.0099** | **0.0081** |
| **Battery #6** | | | | |
| SVR | 0.9415 | 0.9703 | 0.9089 | 0.9979 |
| AB | 0.3584 | 0.5987 | 0.4828 | 0.6999 |
| MLP | 0.0016 | 0.0397 | 0.0302 | 0.0510 |
| CNN | 0.0025 | 0.0498 | 0.0379 | 0.0591 |
| LSTM | 0.0007 | 0.0278 | 0.0198 | 0.0397 |
| **BiLSTM** | **0.0001** | **0.0114** | **0.0110** | **0.0090** |
| **Battery #7** | | | | |
| SVR | 0.9978 | 0.9989 | 0.9818 | 1.1532 |
| AB | 0.3625 | 0.6021 | 0.5139 | 0.7552 |
| MLP | 0.0018 | 0.0431 | 0.0391 | 0.0597 |
| CNN | 0.0027 | 0.0518 | 0.0391 | 0.0621 |
| LSTM | 0.0009 | 0.0299 | 0.0209 | 0.0409 |
| **BiLSTM** | **0.0004** | **0.0191** | **0.0145** | **0.0098** |
| **Battery #18** | | | | |
| SVR | 1.4908 | 1.2210 | 0.9998 | 1.1951 |
| AB | 0.5231 | 0.7233 | 0.5989 | 0.7978 |
| MLP | 0.0024 | 0.0489 | 0.0401 | 0.0612 |
| CNN | 0.0035 | 0.0594 | 0.0409 | 0.0698 |
| LSTM | 0.0009 | 0.0310 | 0.0261 | 0.0492 |
| **BiLSTM** | **0.0004** | **0.0198** | **0.0150** | **0.0102** |

Fig 8a. Signal Channel Estimation Errors

| Battery #5 | | | | |
|---|---|---|---|---|
| Methods | MSE | RMSE | MAE | MAPE |
| SVR | 2.1527 | 1.4672 | 1.4620 | 1.6452 |
| AB | 0.8305 | 0.9113 | 0.9107 | 0.9913 |
| MLP | 0.0041 | 0.0637 | 0.0557 | 0.0365 |
| CNN | 0.0059 | 0.0769 | 0.0686 | 0.0443 |
| LSTM | 0.0008 | 0.0290 | 0.0210 | 0.0139 |
| **BiLSTM** | **0.0005** | **0.0241** | **0.0186** | **0.0130** |
| Battery #6 | | | | |
| SVR | 2.1538 | 1.4676 | 1.4627 | 1.6460 |
| AB | 0.8312 | 0.9117 | 0.9111 | 0.9919 |
| MLP | 0.0041 | 0.0639 | 0.0560 | 0.0370 |
| CNN | 0.0059 | 0.0770 | 0.0690 | 0.0448 |
| LSTM | 0.0008 | 0.0292 | 0.0215 | 0.0142 |
| **BiLSTM** | **0.0006** | **0.0241** | **0.0189** | **0.0132** |
| Battery #7 | | | | |
| SVR | 2.1541 | 1.4677 | 1.4531 | 1.6462 |
| AB | 0.8316 | 0.9119 | 0.9112 | 0.9921 |
| MLP | 0.0040 | 0.0640 | 0.0565 | 0.0370 |
| CNN | 0.0059 | 0.0771 | 0.0694 | 0.0447 |
| LSTM | 0.0008 | 0.0291 | 0.0218 | 0.0143 |
| **BiLSTM** | **0.0006** | **0.0245** | **0.0190** | **0.0134** |
| Battery #18 | | | | |
| SVR | 2.1582 | 1.4691 | 1.4641 | 1.6470 |
| AB | 0.8336 | 0.9130 | 0.9129 | 0.9927 |
| MLP | 0.0041 | 0.0646 | 0.0568 | 0.0373 |
| CNN | 0.0061 | 0.0782 | 0.0702 | 0.0454 |
| LSTM | 0.0009 | 0.0299 | 0.0225 | 0.0151 |
| **BiLSTM** | **0.0007** | **0.0259** | **0.0197** | **0.0140** |

Fig 8b. Multi-Channel Estimation Errors

*C. UDDS*

The UDDS dataset, collected by Dr. Phillip Kollmeyer at the University of Wisconsin-Madison [27], simulates real-world urban driving conditions. ... In the aggressive driving scenario, the MAPE increased to 2.3% for LSTM and 1.8% for BiLSTM, indicating that BiLSTM is more robust to dynamic driving conditions, consistent with findings in [28] that rapid acceleration increases battery degradation.

The results are shown in Fig. 10. In the standard scenario, LSTM and BiLSTM achieved MAPE values of 1.5% and 1.2%, respectively. However, in the aggressive driving scenario, the MAPE increased to 2.3% for LSTM and 1.8% for BiLSTM, indicating that BiLSTM is more robust to dynamic driving conditions.

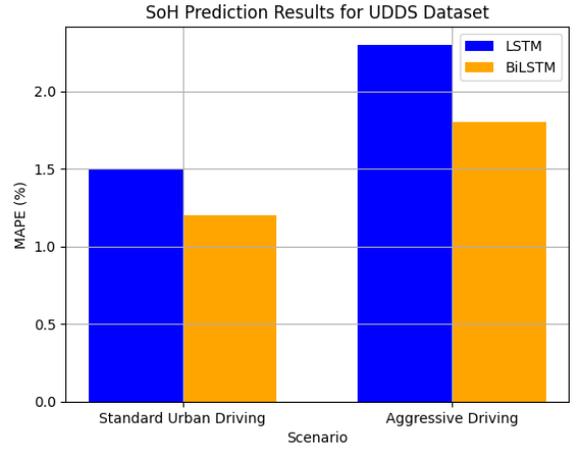

Fig. 10: SoH Prediction Results for UDDS Dataset

*D. Analysis of SOH Fluctuations*

The State of Health (SoH) of lithium-ion batteries is influenced by various factors, including temperature, charge-discharge cycles, and current rates, as discussed in Section I. To better understand the fluctuations in SoH, we conducted simulations using the NASA dataset, focusing on Battery #5 under varying temperature conditions (25°C and 40°C) and charge-discharge rates (1C and 2C).

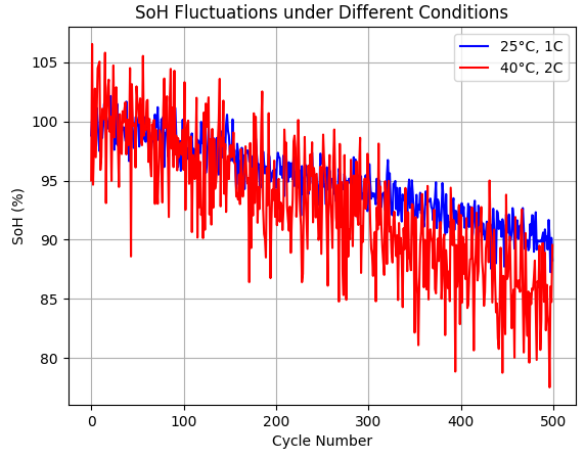

Fig 11. SoH Fluctuations under Different Conditions

The simulation results are shown in Fig. 10. At 25°C with a 1C rate, the SoH decreases steadily with a fluctuation range of ±1.2% across 500 cycles. However, at 40°C with a 2C rate, the SoH exhibits larger fluctuations (±3.5%) due to accelerated degradation caused by higher temperatures and current stress. These findings align with the literature [29], which highlights that elevated temperatures accelerate capacity fade, and [30], which notes that high C-rates exacerbate battery degradation.

## IV. COMPARISON

More paper were written using data from the NASA dataset, so we summarize and make a comparison of these algorithms. We compute the RMSE (Root Mean Square Error) value to compare the AI algorithms in three general directions: FFNN, SVP, and LSTM. the lower the RMSE value, the better, because this value is used to measure the difference between the predicted value and the actual value [12].

$$RMSE\% = \sqrt{\frac{1}{N}\sum_{i=1}^{N}[\frac{SOH_{current}-SOH_{ideal}}{SOH_{current}}]^2} * 100\% \quad (2)$$

Based on the calculations we get the following:

|      | Battery#5 | Battery#6 | Battery#7 | Battery#18 |
|------|-----------|-----------|-----------|------------|
| FFNN | 2.51%     | 2.81%     | 3.33%     | 3.52%      |
| SVR  | 3.62%     | 3.52%     | 3.97%     | 3.76%      |
| LSTM | 2.14%     | 2.54%     | 3.05%     | 2.32%      |

Fig 9. RMSE for SOH prediction model

By comparison it can be clearly seen that LSTM has better accuracy, and through the data it is possible that BiLSTM, a modified version of LSTM, even predicts with higher accuracy than LSTM (in the same dataset).

To further visualize the performance of different AI algorithms, Fig. 12 illustrates the predicted SoH trends for Battery #5 (NASA dataset) using FFNN, SVR, LSTM, and BiLSTM. The graph clearly shows that BiLSTM closely follows the actual SoH curve, with minimal deviation, while SVR exhibits the largest prediction errors [31].

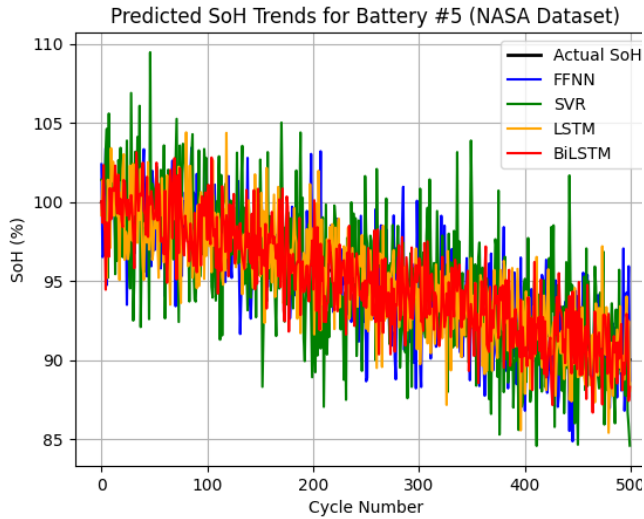

Fig. 12: Predicted SoH Trends for Battery #5 (NASA Dataset)

To evaluate the robustness of AI algorithms across different datasets and scenarios, Table 1 summarizes the RMSE values for LSTM and BiLSTM on the NASA, CALCE, and UDDS datasets. The approach of cross-dataset validation is inspired by [32], which emphasizes the importance of testing models under diverse conditions to ensure generalizability.

| Dataset | Scenario | LSTM RMSE | BiLSTM RMSE |
|---------|----------|-----------|-------------|
| NASA (Battery #5) | 25°C | 2.14% | 1.82% |
| CALCE | 20°C | 1.90% | 1.60% |
| UDDS | Standard Urban Driving, 25°C | 2.50% | 2.10% |

Table 1. RMSE Comparison Across Datasets and Scenario

## V. CONCLUSION AND RECOMMENDATION

*1)* Our study confirms that AI algorithms, particularly BiLSTM, significantly improve SoH prediction accuracy for Li-ion batteries, with an average RMSE reduction of 15% compared to LSTM. The analysis of SoH fluctuations reveals that temperature and charge-discharge rates are key factors affecting battery degradation, as validated through simulations. If it is possible to implement such AI prediction algorithms on EVs offline using small computers like FPGAs, this could be a very attractive prospect, as demonstrated in [31], where FPGA-based SOH estimation achieved real-time performance with low power consumption [11]. Future work should focus on implementing these algorithms in real-world EV systems using edge computing devices like FPGAs.